# NILE: Fast Natural Language Processing for Electronic Health Records

Sheng Yu[1,2,3,*], Tianrun Cai[4,5], Tianxi Cai[6,7,5]

[1]*Center for Statistical Science, Tsinghua University, Beijing, China;*

[2]*Department of Industrial Engineering, Tsinghua University, Beijing, China;*

[3]*Institute for Data Science, Tsinghua University, Beijing, China;*

[4]*Brigham and Women's Hospital, Boston, MA, USA;*

[5]*VA Boston Healthcare System, Boston, MA, USA;*

[6]*Department of Biostatistics, Harvard T.H. Chan School of Public Health, Boston, MA, USA;*

[7]*Department of Biomedical Informatics, Harvard Medical School, Boston, MA, USA;*

* Correspondence to:

Sheng Yu
Weiqinglou Rm 209
Center for Statistical Science
Tsinghua University
Beijing, 100084, China
Email: syu@tsinghua.edu.cn
Tel: +86-10-62783842

**ABSTRACT**

**Objective:** Narrative text in Electronic health records (EHR) contain rich information for medical and data science studies. This paper introduces the design and performance of Narrative Information Linear Extraction (NILE), a natural language processing (NLP) package for EHR analysis that we share with the medical informatics community.

**Methods:** NILE uses a modified prefix-tree search algorithm for named entity recognition, which can detect prefix and suffix sharing. The semantic analyses are implemented as rule-based finite state machines. Analyses include negation, location, modification, family history, and ignoring.

**Result:** The processing speed of NILE is hundreds to thousands times faster than existing NLP software for medical text. The accuracy of presence analysis of NILE is on par with the best performing models on the 2010 i2b2/VA NLP challenge data.

**Conclusion:** The speed, accuracy, and being able to operate via API make NILE a valuable addition to the NLP software for medical informatics and data science.

# INTRODUCTION

Electronic health records (EHR) are a rich source of clinical information that can substantially support biomedical research and healthcare improvement[1,2]. In addition to the structured data such as lab values and billing codes, EHR consist of clinical narratives recorded by healthcare providers that typically contain the majority of the encounter details. The unstructured nature of the narrative data necessitates using natural language processing (NLP) technologies to extract and standardize the information for analysis. Medical language processing (MLP) systems are specially designed NLP systems for clinical narratives to handle their distinct syntax and to extract domain-specific semantics. They are the upstream infrastructure that extract data for researches and applications ranging from EHR-based phenotyping[3–8] to clinical decision support systems[9,10]. Thus, the efficiency, accuracy, and flexibility of MLP is critically important. NILE is an MLP system that has been a backbone of our EHR data processing capability and has been used in a variety of research projects.[11–18] We have released NILE to the research community, and in this paper we introduce its design and performance.

A complete MLP workflow typically involves two major steps: named entity recognition and semantic analysis. Named entity recognition (NER) is the process of identifying mentions of concepts in the text, such as diseases, symptoms, and medications. One commonly used approach for NER is "chunk-and-match". Chunking is the process to identify phrases of interest (usually noun phrases). Many MLP programs, including HITEx[19], cTAKES[20], MetaMap[21], and KnowledgeMap[22], do part-of-speech (POS) tagging before chunking, and use the POS to determine the phrase boundaries. CAPIS[23] uses a simple heuristic to find phrase boundaries. LEXIMER[24] does phrase separation via punctuation and entropy reduction. An extension of CLARIT[25] also employs corpus statistics for the identification of phrases. After chunking, various term identification methods are used to match the phrases against their dictionaries to see if they are recognizable. Another convenient tool for simple term identification is regular expressions. NegEx and ConText[26,27] use regular expressions to identify cues for semantic analysis. HITEx also has a pipeline that uses regular expressions to identify medications from discharge notes. However, simple implementations of regular expressions search for dictionary terms one by one. As a result, the processing speed grows linearly with the size of the dictionary. Therefore, regular expressions are not suitable for EHR processing using a large dictionary like the Unified Medical Language System (UMLS)[28]. An efficient alternative to regular expression based methods is tree-based search algorithms which tend to be extremely fast in speed. The Aho-Corasick algorithm[29] used in MedTagger[30] is an example of such algorithms. NILE also uses tree-based search and is capable of handling prefix and suffix sharing that are common in EHR.

Semantic analysis usually uses the output of the NER step as input and annotates them by the meaning in the text. Common analyses for named entities (symptoms, medications, etc.) include whether they are present/used, conditional, or whether they are hypothetical or part of either the past medical history or family history. Common analyses for attributes of a condition or symptom include determining its location, severity, and chronicity. There are other specialized analyses, such as determining the smoking state of the patient[31,32], value extraction from semi-structured EHR[33,34], extraction of drug and food allergies[35] and medication information[36]. The analysis can be based on rules or be data-driven using machine learning. A number of famous systems, including MedLEE[37], NegEx[26], and ConText[27], use rules based on patterns of cue words. NILE is also rule-based for robustness and efficiency. When annotated sentences are available, machine learning is another good option. The best contenders in the 2010 i2b2/VA challenge unanimously used machine learning for semantic analysis, employing support vector machines as well as other techniques, such as conditional random fields and logistic regressions.[38–45]

There are several reasons for why NILE can be a valuable addition to existing publicly available software such as cTAKES and MetaMap. First, instead of being a software operated via a graphical user interface, NILE takes the form of a Java package. EHR are stored in databases with different schemes, and notes are usually written in diverse formats and templates. So, note parsing usually require customization. A package therefore can be more appealing to data scientists who want to easily write a piece of code to command how to query the EHR database, how to segment the notes, when and where the NLP should kick in, and how the results should be formatted. Second, the UIMA framework[46] adopted by cTAKES and other software is very hard for non-NLP professionals and requires significant time to learn. In contrast, the concepts and APIs used in NILE are easily understood by people who know object-oriented programming. Finally, NILE is extremely fast. Our tests show that NILE is by several orders of magnitude faster than existing software, which is critical to analyzing large EHR databases. At Partners HealthCare and Veterans Affairs, we have used NILE to analyze billions of notes. NILE is currently provided in two forms – a Java package operated via APIs, and a Windows executable incorporating additional features such as multi-threading and database access that can be operated by people without programming background. Download is available at http://celehs.hms.harvard.edu/software/NILE/NILE_main.html.

# DESIGNS AND FUNCTIONS

NILE stands for the Narrative Information Linear Extraction system. The system is called "linear" because NILE's NER and semantic analyzers work by sweeping through the sentence from left to right, allowing NILE to be extremely fast. Medical note processing with NILE typically involves the following steps: 1. Initiate NILE and load the dictionary; 2. Read a note and preprocess it to identify paragraphs that need NLP analysis; 3. Give a paragraph to NILE, which will segment the paragraph to sentences and tokenize them; 4. Apply NER to each sentence; 5. Apply semantic analyses; 6. Write the result to the output device. Step 2 requires the user to understand the target notes and customize for them. Steps 3-5 are packed in a single command for simplicity. Step 6 allows the user to decide what parts of the result are needed and to customize the output format. The rest of this section introduces concepts and designs of NILE's NER and semantic analysis components.

## Named Entity Recognition

The dictionary entries in NILE are called Phrases, which are objects that contain not only the text, but also their semantic roles and codes. There are broadly two kinds of Phrases: 1) grammatical words and meaning cues that we have collected and included as part of the NILE package, and 2) medical terms that need to be provided by the user. For example, the user can load the dictionary with terms extracted from the UMLS. Each Phrase has exactly one semantic role that tells the semantic analyzers how to use the phrase. Currently there are 29 predefined semantic roles, 26 of which are for grammatical words and meaning cues, and 3 (observation, modifier, and location) are for medical terms. When modification information is not needed, which is a common case in note parsing, the user only needs to provide observation terms, which are typically nouns or noun phrases include diseases, symptoms, findings, medications, procedures, etc. A code is used as a concept ID for the Phrase. Typically, if one uses terms extracted from the UMLS, then the Concept Unique Identifier (CUI) will naturally be the code. A Phrase is allowed to have multiple codes. This design for one reason is to reflect the fact that many terms have coding ambiguities. For instance, UMLS concepts C0010054, C0010068, and C1956346 are very close and have common terms "coronary artery disease" and "coronary heart disease". Therefore, when a term with ambiguous codes is identified, the user may wish to output all the codes. Another use case of the multiple codes is to reflect the concept hierarchy. For example, we can find from the UMLS concept relation table that Tylenol (C0699142) is a tradename of Paracetamol (C0000970) and that Paracetamol is a kind of Analgesics (C0002771). Thus, we can let the entry "Tylenol" have all three codes, and when "Tylenol" is identified in a note, all three codes are written to the output. This facilitates concept aggregation and allows one to do flexible feature selection, as seen in automated feature selection for high-throughput phenotyping.[12,13]

The dictionary of NILE uses a prefix tree data structure. Figure 1 illustrates a prefix tree with five terms: "heart", "heart failure", "heart attack", "coronary angiogram", and "coronary artery". If a term ends at a node, then the node contains the corresponding Phrase object; otherwise that field is null. Each node uses a hash table to connect to its children via the next token (word) in the term. The hash tables make the search speed insensitive to the number of entries in the dictionary. The matching algorithm reads a sentence as a series of tokens, and matches the longest phrase from left to right. For instance, when NILE sees "patient had a heart attack in 2006…" it identifies "heart attack" rather than "heart", and goes on to find the next identifiable phrase starting from "in". We extended the basic prefix tree search algorithm to recognize prefix and suffix sharing. Supplementary Material Algorithm 1 is the pseudo code that returns the longest phrase starting at the i-th token if the first argument `node` is the root. The biggest if-block extends the basic algorithm to recognize suffix sharing, so that "mediastinal, hilar, or axillary lymphadenopathy" can be recognized as "mediastinal lymphadenopathy, hilar lymphadenopathy, or axillary lymphadenopathy". Prefix sharing is handled in the following way: if the previous identified term is at least two words long, keep the first word as prefix; if matching at the next position fails, try again with `match(root.child(prefix), tokens, i)`. The prefix is reset to empty if the match fails again or returns the prefix itself. With both prefix and suffix sharing recognition, we can now interpret "right upper, middle, and lower lobes" as "right upper lobe, right middle lobe, and right lower lobe".

NILE's NER is case sensitive. In general situations, we encourage the user to convert both the dictionary terms and the input sentences to lowercase. The cases in clinical notes can be very unpredictable – some notes or paragraphs are written entirely in uppercase. So, while lowercasing everything can bring ambiguity to some acronyms, we find it a good trade-off in general. Another thing to note is lemmatization, which is a common step in NER, typically achieved via the LVG lexical tools.[47] In the design of NILE, we have avoided lemmatization to keep the matching algorithm pure and minimal; and to match plural forms, we expect the user to populate the dictionary with plural forms using the LVG.

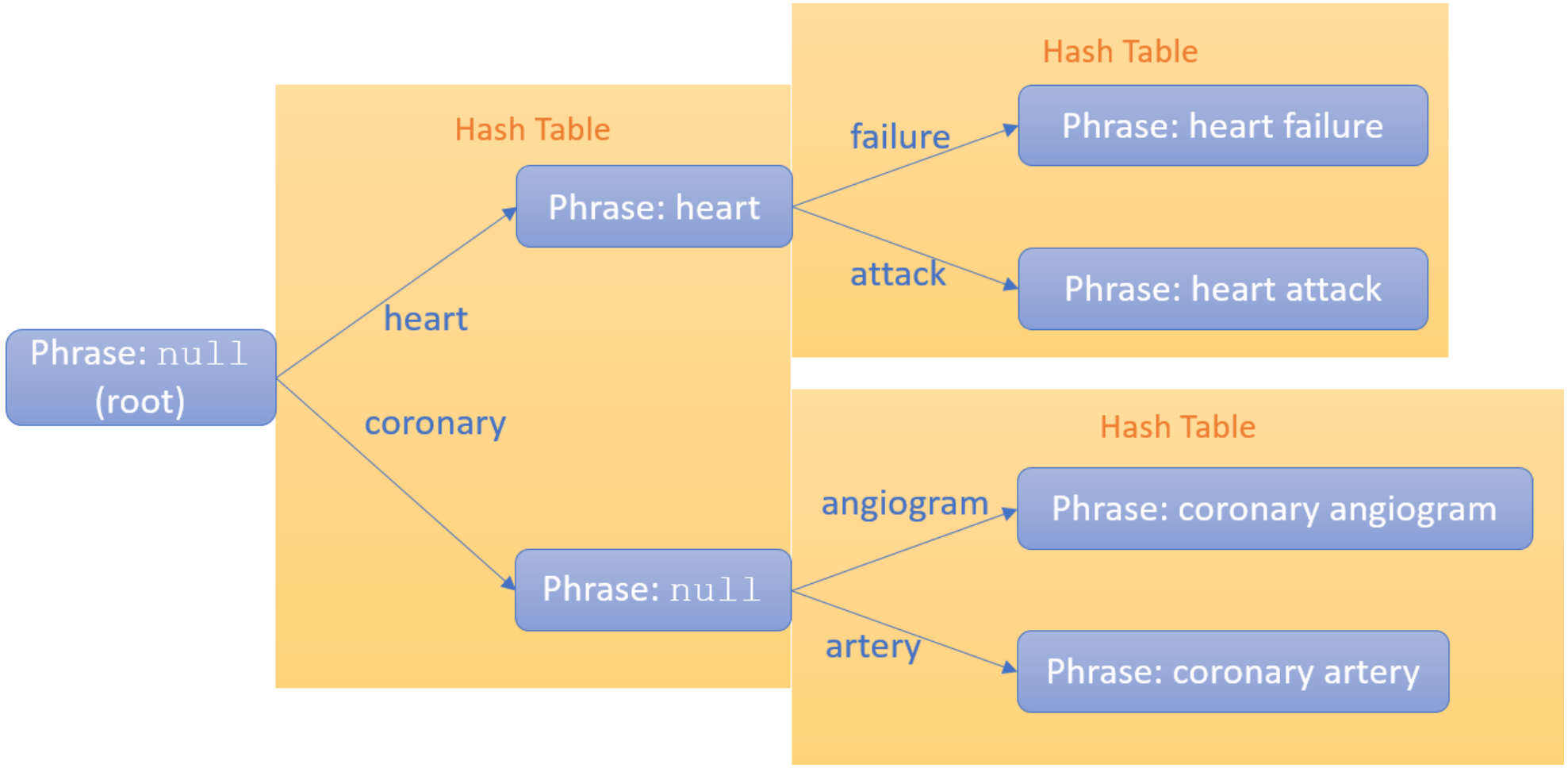

Figure 1: Dictionary data structure

## Semantic Analysis

The output of the NER step is a list of Semantic Objects, which are essentially Phrases packed with additional offset information. Words and phrases not in the dictionary are not kept in the NER output. The semantic analysis step involves a sequence of analyzers with a common API that uses `List<SemanticObject>` as both input and output. Each analyzer may modify attributes of the Semantic Objects (e.g., presence and family history), attach Semantic Objects to other ones (modification), or remove Semantic Objects from the list. The analyzers are implemented as finite state machines that look at the Semantic Objects in the list sequentially, modify their states depending on the current states and what objects they see, and decide their actions on each Semantic Object. Each analyzer is designed to be simple, so an analysis (e.g., presence analysis and location analysis) may comprise multiple analyzers. Implemented analyses include presence, location, modification (other than location), family history, and an ignoring analyzer that will remove an identified observation object from the output if it does not express presence or absence. For example, "patient came to rule out rheumatoid arthritis" and "Breast Cancer Center" do not express presence or absence of "rheumatoid arthritis" or "breast cancer", so these observation objects will be removed, and the user will not see them in the output. In the following, we introduce presence and location analysis in more detail.

### Presence Analysis

The presence analysis (also known as negation analysis) analyzes whether a problem is present, a drug is prescribed/used, a test is taken, etc., and assigns values YES, NO, or MAYBE. The analysis is performed in two steps (two analyzers), starting with preprocessing and followed by presence tagging. The preprocessing checks if consecutive meaning cues or grammatical words can form a new cue. For example, originally in the dictionary, "found" is a past participle for confirmation. When the analyzer detects a sequence in the left column of Table 1, it replaces the Semantic Objects with a new object with the role in the right column. The analyzer only looks at the semantic roles, so that the rules can generalize. For example, if "found" is replaced by "identified", or if "isn't" is replaced by "weren't", they would appear the same to the analyzer.

Table 1: Combined effects of semantic roles.

| **Combination** | **Semantic Role** |
|---|---|
| found (used alone) | Confirmation |
| not + found | backward negation |
| have + found | Confirmation |
| haven' t + found | Negation |
| have + been + found | backward confirmation |
| haven't + been + found | backward negation |
| is + found | backward confirmation |
| isn't + found | backward negation |

After the preprocessing, the presence tagger iterates through the Semantic Objects. When the tagger sees an observation object, it applies the current presence state to it and puts it in the cache. Depending on the semantic role or the role transition of the observed objects, the tagger may change the presence state, change the attributes of the observation objects in the cache (when seeing backward cues, e.g., "was not found"), or empty the cache. Figure 2 provides an illustration. The analyzer initiates with presence state YES and starts from the first semantic object. Since "no" is a negation cue, the presence state changes to NO. The analyzer then proceeds to "evidence", which is a confirmation cue. In the current implementation of the analyzer, negation always suppresses confirmation, so the state remains as NO. The analyzer does nothing when it sees the preposition "of" and goes on. Finally, when the analyzer sees the observation object "pulmonary embolism", it applies the current state NO to the object's presence attribute.

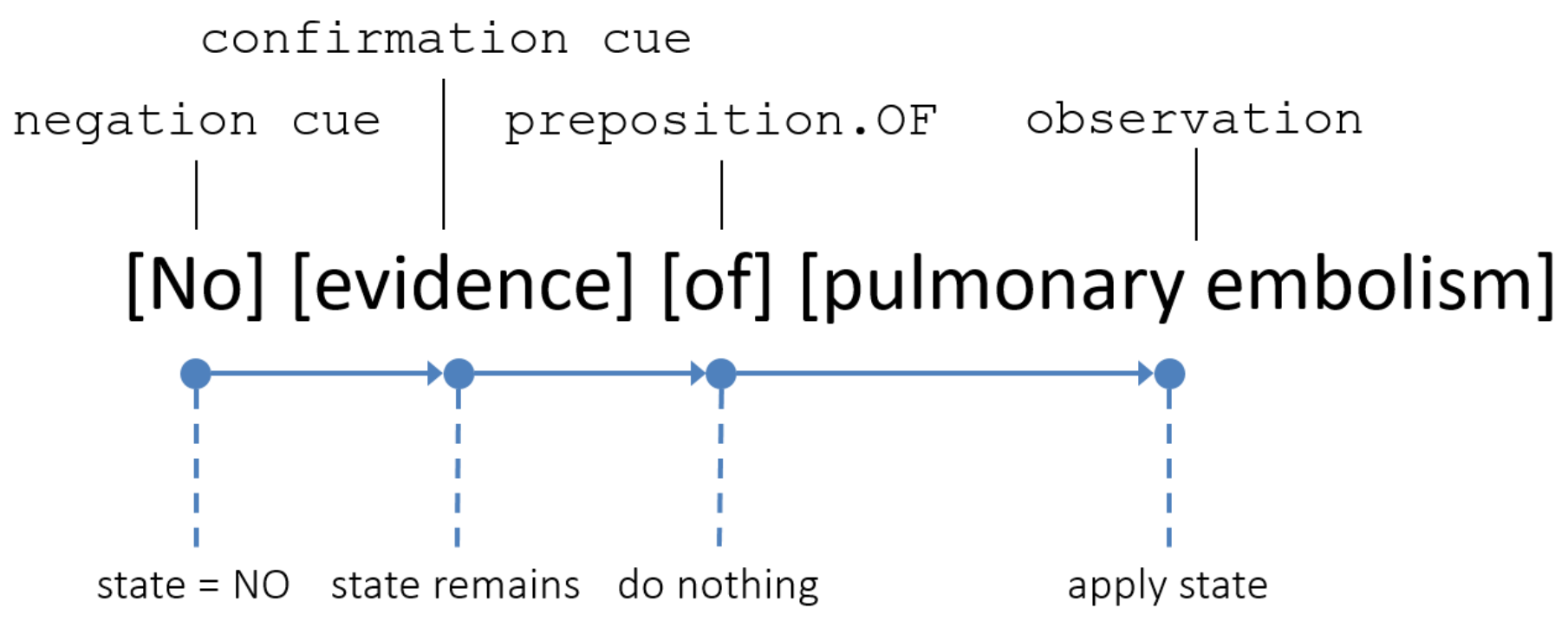

Figure 2: Illustration of the presence analysis.

With enough granularity of semantic roles and rules, the presence analyzers can correctly classify the presence most of the times, including subtle differences as shown in Table 2, which are usually too hard for existing software. We have also added support for presence analysis from other analyzers. For example, words like “change”, “increasing”, and “worsening” are given a semantic role that represents change, and an analyzer that precedes presence analysis detects them and their targets, so that later in “no change in pleural effusion”, the negation will only apply to “change” but not to “pleural effusion”.

Table 2: Handling subtle differences in presence analysis.

| **Example Sentence** | **Return** |
|---|---|
| No filling defects are seen to suggest pulmonary embolism. | filling defects: NO<br>pulmonary embolism: NO |
| No filling defects are seen, suggesting pulmonary embolism. | filling defects: NO<br>pulmonary embolism: YES |
| No filling defects are seen and it suggests pulmonary embolism. | filling defects: NO<br>pulmonary embolism: YES |
| No filling defects that suggest pulmonary embolism are seen. | filling defects: NO<br>pulmonary embolism: NO |
| No filling defects are seen, which suggests pulmonary embolism. | filling defects: NO<br>pulmonary embolism: YES |

**Location Analysis**

Identification of problem locations is another useful analysis in MLP. NILE can handle sophisticated sentences and do nesting analysis for anatomical locations, which is an important feature that is missing in other existing MLP programs.

Descriptions of anatomical locations usually involve nested modifications that can be expressed in very diverse ways. The simplest patterns of nested modifications are "location A location B" (e.g., "left upper lobe arteries") and "location A of/in/within/... location B" (e.g., "arteries of the left upper lobe"). The analyzer iterates through the sequence of Semantic Objects, and treats the former pattern as "`location B (location A)`" and the latter as "`location A (location B)`", where objects in the parentheses modify the object preceding the parentheses. In the implementation, the object inside of the parentheses is removed from the Semantic Object list and its reference is added to the modifier list of the object preceding the parentheses. These patterns can be chained. For instance, "location A location B location C" is interpreted as "`location C (location B (location A))`". The same patterns are used to assign location objects to the observation objects that they modify, except that observation objects cannot be modifiers. For example, "location A fact B" means "`fact B (location A)`", and "fact A in location B location C" means "`fact A (location C (location B))`".

The above patterns are simple because they involve a single nesting chain. The analysis becomes difficult when the modification structure has branches (such as two modifiers joined by "and"). The following is an example from a computed tomography pulmonary angiography (CTPA) report:

"*There are segmental and subsegmental filling defects in the right upper lobe, superior segment of the right lower lobe, and subsegmental filling defect in the anterolateral segment of the left lower lobe pulmonary arteries.*"

This sentence includes multiple observation objects, each having multiple location modifiers, whose modification structures contain both nesting and branches. The location analyzer checks the semantic roles of the current and the previous object as well as the nesting state (forward/backward) to decide whether to nest or to branch. For the above sentence, using a custom CTPA dictionary[11], NILE returned:

```
filling defects: YES (right upper lobe; superior segment (right lower lobe); segmental; subsegmental),
filling defect: YES (segment (pulmonary arteries (left lower lobe)); subsegmental),
```

which is the same as the intended meaning, except that "anterolateral segment" was not in the dictionary.

# EVALUATION

We provide speed and accuracy evaluations for NILE. The computer for benchmarking the speed has an AMD Ryzen 7 2700X CPU and 32 GB RAM, running Java 1.8.0_191. The data set used for testing the presence analysis accuracy is the 2010 NLP i2b2/VA challenge data, available at https://i2b2.org/NLP/DataSets.

## Speed

We compare NILE with cTAKES 4.0.0 and MedTagger 1.0.1 to process 10,330 CTPA reports. cTAKES is a popular open source MLP system that uses the "chunk-and-match" approach for NER and uses machine learning based modules for semantic analysis. MedTagger mainly focuses on NER and uses a prefix-tree search algorithm, so it is expected to be faster than cTAKES, but it does not have the ability to detect prefix and suffix sharing like NILE. Both cTAKES and MedTagger were run with their default dictionaries and recommended memory allocation (3 GB for cTAKES, 2 GB for MedTagger). We tested NILE (single-threaded) with two sets of dictionaries: a small dictionary with about 1000 terms that we curated for a pulmonary embolism study[11], and a large dictionary with an additional 0.9 million terms arbitrarily drawn from the UMLS. Table 3 shows the average processing time for the 10,330 CTPA reports, where the time for cTAKES and MedTagger are the average of two runs, and the time for NILE for both situations are the average of 10 runs.

Table 3: Average processing time for 10,330 CTPA reports.

| Software/setting | Processing time |
|---|---|
| **cTAKES 4.0.0** | 4305.55 sec |
| **MedTagger 1.0.1** | 816.17 sec |
| **NILE (small dictionary)** | 1.84 sec |
| **NILE (large dictionary)** | 2.18 sec |

## Accuracy of Presence Analysis in Discharge Summaries

The 2010 i2b2/VA NLP challenge dataset contains discharge summaries annotated with the problem terms of interest and a corresponding label for their presence with six possible values: “present”, “absent”, “possible”, “hypothetical”, “conditional”, and “associated_with_someone_else”. To accommodate for these annotations, we customized NILE and combined several analyzers to jointly classify the presence, with the exception of “conditional”, as we did not have an analyzer for it. The dataset contains 170 reports for training and 256 reports for testing, which are fewer than the reported size in the 2010 challenge (394 and 477 reports for training and testing, respectively).[45] NILE’s analyzers are rule-based and do not need training. Instead, we used the training data to understand the annotator’s logic and reassigned the semantic roles of some of the cue words accordingly. This customized version is released together with the standard version of NILE. We encourage using the standard version because it suits more general scenarios (see Supplementary Material for details).

Table 4 shows the F scores of NILE for classifying the presence state on the test data, with comparison to the reported scores of deBruijn[38] (system 2.3) and MITRE[39], which ranked 1st and 2nd, respectively, in the 2010 i2b2/VA challenge. The confusion matrices are given in the Supplementary Material.

Table 4: F scores for classifying the presence state on the 2010 i2b2/VA challenge data.

| | NILE | deBruijn | MITRE |
|---|---|---|---|
| **absent** | 0.934 | 0.929 | 0.94 |
| **someone_else** | 0.954 | 0.899 | 0.87 |
| **conditional** | -- | 0.355 | 0.42 |
| **hypothetical** | 0.860 | 0.883 | 0.89 |
| **possible** | 0.627 | 0.627 | 0.63 |
| **present** | 0.954 | 0.952 | 0.96 |

# DISCUSSION

NILE is a time tested software initially developed in 2012 and has been used in many projects over the years[11–18]. The most important merit of NILE is that it is significantly faster than other publicly available platforms while maintaining a comparable accuracy. As Table 3 shows, NILE is 2000 times faster than cTAKES and 400 times faster than MedTagger, and the dictionary size minimally affect the processing speed. As the amount of EHR data available for analysis keeps growing, a large processing bandwidth has become a critical quality to have for medical data science. The speed of NILE has given us the capability to process very large EHR data sets, such as performing phenome-wide association studies (PheWAS) using million-patient scale EHR data from the VA MVP[48]. In practice, we find that the majority of processing time is spent on the I/O (reading database and writing the result) rather than on the NLP, so we did not incorporate multi-threading for NILE in the core package. However, a Windows executable version with multi-threading capability is available for download.

Choosing rule-based semantic analysis over machine learning is another design decision. While NLP improves rapidly with the current research of deep learning, the availability of large-scale annotated data is a major obstacle to widely using machine learning. In comparison, the development of rule-based semantic analyzers is not limited by the availability of annotated data, and their performances are generally acceptable. Table 4 shows that the accuracy of NILE's presence analysis is on par with the top performing machine learning models from the i2b2/VA challenge. As detailed in the Supplementary Material, NILE's presence analyzer have different annotation labels and logic from that used in the challenge. Being able to customize NILE for this benchmark also reflects NILE's flexibility from being a package operated via API, which can be appealing to data scientists in medical informatics.

From the experience of using NILE, we have found a few limitations of it that are worth noting. One limitation is that the values of the presence state (YES, NO, or MAYBE) do not fully cover the possibilities, especially the time dimension. For example, in "the patient has ceased smoking" and "the symptom has disappeared", simply assigning YES or NO cannot fully record the information from the sentences. Another limitation is term sense disambiguation. Acronyms often have multiple possible meanings. For example, "CHD" could represent congenital heart disease, coronary heart disease, or child/children; "DR" could be doctor, drive, delayed release, or diabetic retinopathy; "RA" could be rheumatoid arthritis or room air (for testing oxygen saturation). NILE does not do disambiguation analysis. However, the problem can be addressed outside of NILE, e.g., using distributional semantics[49,50].

# CONCLUSION

In this paper, we introduced the design and performance of NILE, an NLP package for EHR analysis that we share with the medical informatics community. NILE is hundreds to thousands times faster than existing software, with accuracy on par with the best performing models in the 2010 i2b2/VA challenge. These qualities along with being able to program via API make it an appealing addition to the existing NLP software for data scientists in this field.

**Funding**: None.

**Competing Interests**: None.

**Contributorship**: Sheng Yu developed the original NILE as a Java package. Tianrun Cai developed the Windows executable version with additional features including multi-threading and database access. Tianxi Cai provided guidance on the design and testing of the NILE package. All authors contributed to evaluating the performance of NILE in analyzing EHR data and drafting and critically revising this manuscript.

# Supplementary Material

# NILE: Fast Natural Language Processing for Electronic Health Records

Sheng Yu[1,2,3,*], Tianrun Cai[4,5], Tianxi Cai[6,7,5]

*[1]Center for Statistical Science, Tsinghua University, Beijing, China;*

*[2]Department of Industrial Engineering, Tsinghua University, Beijing, China;*

*[3]Institute for Data Science, Tsinghua University, Beijing, China;*

*[4]Brigham and Women's Hospital, Boston, MA, USA;*

*[5]VA Boston Healthcare System, Boston, MA, USA;*

*[6]Department of Biostatistics, Harvard T.H. Chan School of Public Health, Boston, MA, USA;*

*[7]Department of Biomedical Informatics, Harvard Medical School, Boston, MA, USA;*

* Correspondence to:

Sheng Yu
Weiqinglou Rm 209
Center for Statistical Science
Tsinghua University
Beijing, 100084, China
Email: syu@tsinghua.edu.cn
Tel: +86-10-62783842

# Prefix tree search with suffix-sharing recognition

```
function match(node, tokens, i) {
  if (node == null) return null
  if (i >= tokens.length) return null
```

```
  p = match(node.child(tokens[i]), tokens, i + 1)
  if (p != null) return p
  if (tokens[i] == "and" OR "or" OR ",")
    k = LIMIT // a constant integer, e.g., 3 or 4
    for (j = i + 2; j < tokens.length AND k > 0; j++, k--)
      if (tokens[j] == "and" OR "or" OR ",")
        k = LIMIT
        continue
      if (node.child(tokens[j]) != null)
        p = match(node, tokens, j)
        break
  if (p != null) return p
  return node.phrase
}
```

Algorithm 1: Prefix tree search with suffix-sharing recognition

# Customization and Results for the 2010 i2b2/VA challenge data

The 2010 i2b2 NLP data set contains clinical narrative texts that have been chunked to sentences, i.e., ideally each line contains one complete sentence. Each sentence has also been tokenized with tokens separated by space. The reference data specifies the problem terms of interest and provides a value for their presence, which could be "present", "absent", "possible", "hypothetical", "conditional", or "associated_with_someone_else". The "present", "absent", and "possible" correspond to the YES, NO, and MAYBE values in our presence analyzer. The value "hypothetical" is given in cases such as "call if you experience shortness of breath". It is similar to our analyzer for ignoring, thus we modified it to detect the hypothetical problems. The value "conditional" is given when the sentence states that a problem can be triggered by certain conditions. We do not have similar analysis, thus we removed them from the test. The value "associated_with_someone_else" means family history most of the times, and we used our family history analyzer for it.

One issue of the reference is that it also considers the meaning of the specified terms. For instance, a term could be "afebrile", meaning "no fever", and the reference value for presence would be "absent". The way NILE works is to assign YES to the presence state to "afebrile" rather than NO to "fever". Thus, to accommodate this problem, we added a postprocessor and forced the presence of a term to be "absent" if it was known to express negation by itself. Another ad hoc customization was the detection of the family history section, and all problem terms in the section were labeled as “associated_with_someone_else”.

The data set includes training data and testing data. Our analyzers are rule-based and do not need training. Instead, we read and test ran with the training data to understand the annotator’s logic and modified our cue words accordingly. For instance, normally we use “evaluate” and “rule out” as cues to ignore a part of a sentence, but in this test we made them speculation cues so that their objects would be labeled as “possible”. The terms that self-expressed negation were also found with the test run.

This customized version of NILE is available upon request.

| | Prediction | | | | | |
|---|---|---|---|---|---|---|
| Reference | absent | someone_else | hypothetical | possible | present | **sensitivity** |
| absent | 1527 | 1 | 3 | 4 | 62 | 0.956 |
| someone_else | 1 | 82 | 1 | 0 | 5 | 0.921 |
| hypothetical | 3 | 0 | 344 | 1 | 34 | 0.901 |
| possible | 15 | 0 | 4 | 195 | 95 | 0.631 |
| present | 77 | 7 | 29 | 77 | 4432 | 0.959 |
| **precision** | 0.941 | 0.911 | 0.903 | 0.704 | 0.958 | |

Table 5: Confusion matrix for the training data set

| | Prediction | | | | | |
|---|---|---|---|---|---|---|
| Reference | absent | someone_else | hypothetical | possible | present | **sensitivity** |
| absent | 2434 | 3 | 5 | 8 | 146 | 0.938 |
| someone_else | 1 | 125 | 0 | 0 | 5 | 0.954 |
| hypothetical | 6 | 0 | 400 | 5 | 34 | 0.899 |
| possible | 21 | 0 | 16 | 366 | 249 | 0.561 |
| present | 154 | 3 | 64 | 136 | 8259 | 0.959 |
| **precision** | 0.930 | 0.954 | 0.825 | 0.711 | 0.950 | |

Table 6: Confusion matrix for the testing data set